\documentclass{article}
\usepackage{hyperref}

\title{The test set for the TransCoder system}

\author{Ernest Davis \\ Dept. of Computer Science \\ New York
University \\ davise@cs.nyu.edu}

\begin{document}
\maketitle
\begin{abstract}
The TransCoder system translates source code between Java, C++, and 
Python 3. The test set that was used to evaluate its quality is missing
important features of Java, including the ability to define and use 
classes and the ability to call user-defined functions other  than 
recursively. Therefore, the accuracy of TransCoder over programs with those
features remains unknown.
\end{abstract}

\section{Unsupervised learning for translating between programming languages}

The TransCoder program (Lachaux et al., 2020) translates source code between 
the three programming languages Java, C++, and Python 3. It was built
using techniques that have been developed for unsupervised machine learning of
translators for underresourced natural language (Lample et al., 2017).
The significance of ``unsupervised'' here is in contrast to earlier 
approaches to machine translation, which have always required
having a corpus of ``bitexts''; that is, text that have been translated
by competent human translators from one language to the other. For 
underresourced languages, the available corpus of bitexts may be too small
to support the automated learning of machine translators. In unsupervised
learning, by contrast, all that is required are unrelated text corpora in
the two languages plus a kernel of corresponding words --- a small bilingual
dictionary. Using ingenious machine-learning techniques, this can be 
bootstrapped to find more numerous and more complex correspondences and thus
construct an effective translation program.

The corresponding procedure applied to translating programs between two different
languages means that one starts with a corpus of programs in each of the two
languages, which of course can be found online in enormous quantity, plus
some basic correspondences. 
For example, the key word ``for'' means very much the 
same thing in all three programming language C++, Java, and Python (and 
many others) and a for-loop in one language
can almost always be translated as a for-loop in the other two. 
Again, starting with these corpora and these basic correspondence, the
machine learning system finds its way to translate programs in general
in one language into the other.

I find it very remarkable, indeed quite counter-intuitive,
that this works at all, in either domain, but good results have obtained
in both. As regards the TransCoder, Roziere et al. (2020) write, in a 
Facebook AI Research blog, 

\begin{quote}
In our evaluations, the model correctly translates more than 90 percent 
of Java functions to C++, 74.8 percent of C++ functions to Java, and 
68.7 percent of functions from Java to Python. 
In comparison, a commercially available tool translates only 61.0 
percent of functions correctly from C++ to Java, and an open source 
translator is accurate for only 38.3 percent of Java functions translated 
into C++.
\end{quote}

The key phrase there, though, is ``in our evaluations''. How was the system
evaluated?  What was the test set?

\section{The Test Set for TransCoder}
The Facebook blog (Roziere et al. 2020) does not mention the test set at all,
and the technical paper (Lachaux et al. 2020) gives only the following short
account:

\begin{quote}
GeeksforGeeks is an online 
platform\footnote{\url{https://practice.geeksforgeeks.org/}} 
with computer science and programming articles. It gathers
many  coding  problems  and  presents  solutions  in  several  
programming  languages.   From  these solutions, we extract a set of 
parallel functions in C++, Java, and Python, to create our validation 
and test sets. These functions not only return the same output, but 
also compute the result with similar algorithm.
\end{quote}

The test set, 
with much other material, has now been published on
github\footnote{\url{https://github.com/facebookresearch/TransCoder}} 
so interested researchers can check it for themselves. 

In supervised machine learning, the standard procedure, for good reasons, is
to divide the labelled corpus randomly into a training set and a test set
(plus, often, a validation set). This certainly cannot be done in unsupervised
machine translation of natural language; 
the training set is a pair of monolingual corpora, but to
evaluate translation, the use of bitexts is unavoidable. This is the model
that has been followed in TransCoder; the corpus of programs from GeeksForGeeks
has been used as a test set because it is a natural corpus of corresponding
programs in different languages. In further studies of translation between
programming languages, it might actually be possible to use the same monolingual 
datasets
using the ``computational accuracy'' metric introduced in (Lachaux et al. 2002);
namely, you take a program in Java, the system translates it into Python,
and you test what fraction of the time the two programs give the same output
on valid inputs. However, this is not what they did, perhaps because they
wanted to compare the validity of computational accuracy with the more commonly
used BLEU score, which necessarily requires bitexts.

The test set consists of six data sets: a validation set and a test set in
each of Java, C++, and Python. I examined the test set of Java with some care;
all the results below are taken from that set. I looked cursorily at the
other files to make sure that they didn't seem to be extremely different
in the respects that I will discuss.

{\bf The bottom line:} Key features of the programming languages are not 
represented
at all in the test set. The evaluation therefore gives no information whatever
about how well the translator handles those features.

Specifically, in the Java test set:

\begin{itemize}
\item
None of the examples involve defining a class. Indeed, the keyword {\tt class}
(used in all three languages)  does
not appear in {\em any\/} of the test or validation sets.
\item
The only dynamic data structures that are used are those defined in a few
standard libraries; and those are not very frequent. In a large fraction of
the examples, the only data types are {\tt int}, {\tt char}, and {\tt String},
and one- or two-dimensional
arrays of {\tt int}, 
\item 
All or almost all the function calls are direct recursive calls of a function
calling itself, or calls to library functions. I did not observe any cases where
one user defined function calls a different one; if there are any, they are
few.
\item The examples tend to be short. In the Python test set, there are 868 
examples\footnote{Lachaux et al. states that the test sets have 852 examples.
For the most part, an example constitutes one line in the text file I examined,
and there were 868 lines. I did not investigate the reason for this
discrepancy of 16. Perhaps
there are 16 cases where an example consists of two function definitions, 
each of which is a line in the text file.}
and a total of 9956 line breaks; that indicates an average of about 11 lines
per example. ``Line of code'' is a less well defined measure in Java; 
but the number of semicolons is a reasonable proxy. The 868 examples
include 8406 semicolons; thus, a similar measure.
\end{itemize}

To further characterize the Java language features used in the test set:
Let us say that the following features, which would normally be taught
in the first month or so of a Java programming class,  are {\em elementary}:
\begin{itemize}
\item The data types {\tt int, int[], int[][], float,  double, bool, char,}
and {\tt String}. (Arrays of {\tt float, double,} or {\tt bool} are extremely
rare.)
\item Basic arithmetic, boolean, and assignment operators, array indexing,
parentheses, and the function {\tt String.charAt()}.
\item The control key words  {\tt if, else, white, repeat, return} and
the use of the open and close curly bracket.
\item {\tt System.out.print} and {\tt System.out.println}.
\item Function definition.
\end{itemize}

Then, of the first 100 examples in the Java test set:\footnote{This count
was made rapidly manually and may well be off by a few, but it is unlikely
to be very far off.} 
\begin{itemize}
\item 45 use only elementary features.\footnote{This is just a characterization
of the language features used; some of the examples were fairly sophisticated 
algorithmically, considering the length of the code.}
\item 14 use elementary features plus some basic functions from the
{\tt Math} package, such as 
{\tt Math.max, Math.min, Math.abs, Math.sqrt, Math.pow}
\item 2 use elementary functions plus recursion.
\item 1 uses elementary functions plus recursion plus functions form the
{\tt Math} package.
\end{itemize}

38 use more sophisticated features. These include:
\begin{itemize}
\item Less common control words, such as {\tt break}, {\tt continue},
{\tt case}, and {\tt switch}. The exception handlers
{\tt try}, and {\tt catch} occur, twice each, further on
in the file. (The frequencies of these
in the file as a whole are shown in table~\ref{tabOccurrenceCount}.)
\item Bitwise operations.

\item Other built-in functions such as {\tt sort, equals} and {\tt clone}

\item
Some functions or constants associated with wrapper classes
such as {\tt Character.isDigit, 
Integer.MaxValue, Arrays.sort, Arrays.binarySearch, Arrays.fill,
Arrays.stream, Integer.toString,} and {\tt Integer.parseInt}.

\item Library classes and associated methods, such as 
{\tt Vector$<$Integer$>$, 
HashMap$<$Integer,Integer$>$, Stack$<$Integer$>$, List$<$Integer$>$
LinkedHashSet$<$Integer$>$, Queue$<$Integer$>$,} and \\
{\tt  StringBuffer}.
\end{itemize}

Any other kinds of features are extremely rare or non-existent in the
test file. (A possible exception here is casting, which occurs at least 
occasionally in the Java file, and which is easy to miss in a quick
manual scan.)

Table~\ref{tabOccurrenceCount}
shows the number of occurrences of various symbols and keywords in the
Java test file.

\begin{table}
\begin{center}
\begin{tabular}{l|r|c|l|r}
{\bf Symbol} & {\bf Occurrences} & \hspace*{0.5in} & {\bf Reserved word} 
& {\bf Occurrences}\\ \hline
Programs & 868 & & for & 1306 \\ \hline
; & 8406 & 
& {\tt if}  & 1401 \\ \hline
\{ & 2428 & 
& {\tt else} & 369\\ \hline
\} & 2427 & 
& {\tt while} & 217 \\ \hline
{[} & 5685 &  
& {\tt repeat} & 4  \\ \hline
{]} & 5685 & 
& {\tt return} & 1116\\ \hline
( & 7073 & 
& {\tt switch} & 3  \\ \hline 
) & 7073 & 
& {\tt case} & 7  \\ \hline 
+ & 1741 & 
& {\tt break} & 79\\ \hline
$-$ & 1915 & 
& {\tt continue} & 31\\ \hline
* & 493  & 
& {\tt try} & 2  \\ \hline
/ & 243 &
& {\tt int} & 4701 \\  \hline
++ & 1503 &
& {\tt double} & 105 \\ \hline
$--$ & 200 &
& {\tt float} & 43 \\ \hline
&  &
& {\tt char} & 100 \\ \hline
&  &
& {\tt bool} & 146 \\ \hline
&  &
& {\tt Integer} & 310 \\ \hline
&  &
& {\tt String} & 268  \\ \hline
& &
& {\tt sort} & 54 \\ \hline
& &
& {\tt equals} & 6 \\ \hline
\end{tabular}
\end{center}
\caption{Occurrences of symbols and reserved words in Java test set}
\label{tabOccurrenceCount}
\end{table}

\section{Features of Java not tested}

The evaluation that has been carried out entirely
omits some of the most critical
and common features of Java programming, most notably the ability to
define classes with owned methods and to use objects in those classes.
Also omitted are abstract classes, interfaces, 
generics (except for standardized uses of library
classes), defining exceptions, dynamic data structures (again except
for library classes), and calling functions/methods non-recursively.

There is good reason to suppose that the problem of these omitted features
is considerably easier than the features that are included in the test tile,
especially the elementary features that constitute a large fraction of the
test file.

Java and C++ are, in fact, extremely similar in how they handle, and 
how they name, the elementary features; and many other languages, including
Python, are quite similar. Many of these features have been more or less
standard in imperative and object-oriented languages since C, in 1970.
For instance, table~\ref{tabComparison} shows the first elementary 
function in the test sets in the three 
languages.\footnote{The title is the one given in the dataset. ``Efficient'' 
is a misnomer; this is an $O(n^{2})$ algorithm to solve a 
problem that has an extremely easy $O(n)$ solution.}

\begin{table}
\begin{verbatim}
EFFICIENTLY COMPUTE SUMS OF DIAGONALS OF A MATRIX

Java 
static void printDiagonalSums(int[ ][ ] mat, int n) { 
   int principal=0 , secondary=0; 
   for (int i=0; i<n; i++) {
      for (int j=0; j<n; j++) { 
         if (i==j) principal+=mat[i][j]; 
         if ((i+j)==(n-1)) secondary+=mat[i][j]; 
        } 
    } 
    System.out.println("Principal_Diagonal: " + principal); 
    System.out.println(" Secondary_Diagonal: " + secondary); 
 }

C++
void printDiagonalSums (int mat[][MAX], int n) { 
    int principal=0 , secondary=0 ; 
    for (int i=0; i<n; i++) { 
       for (int j=0; j<n; j++) { 
         if (i==j) principal+=mat[i][j]; 
         if ((i+j)==(n-1)) secondary+=mat[i][j]; 
      } 
    } 
    cout << "Principal_Diagonal: " << principal << endl; 
    cout << "Secondary_Diagonal: " << secondary << endl ; 
 }

Python 3 
def printDiagonalSums (mat,n): 
    principal = 0 
    secondary = 0  
    for i in range (0,n): 
        for j in range(0,n): 
            if (i== j): 
                principal+=mat[i][j] 
            if ((i+j)==(n-1)) 
                 secondary+=mat [i][j] 
    print("Principal_Diagonal:", principal) 
    print("Secondary_Diagonal:", secondary ) 
\end{verbatim}
\caption{Comparison of an elementary function in Java, C++, and Python}
\label{tabComparison}
\end{table}

The Java and the C++ code are character-for-character identical in lines
2-7. They differ only in the form of the declaration of the function
and the array argument, and in form of the outputting statements. 
The Python code is more different: the {\tt for} statement has a different
form, there are no type declarations, semi-colons are replaced by new lines,
curly brackets are replaced by indentation. But a student who has learned
to write this kind of code in Java or C++ can quickly learn to write
it in the other languages; going from Python to the other two is a little
more demanding but not very much so.

By contrast, there are large and fundamental differences in the ways in which 
the three languages handle (or fail to handle) typing, referencing and 
dereferencing, class hierarchies and inheritance, overloading, dynamic 
and static dispatching, memory management, and other deep language 
features. Translating a program that uses these in sophisticated ways
from one language to another
is by no means a cookie-cutter process; it can require careful inspection
and analysis and significant redesign.

How well would TransCoder do on code that includes these untested features?  
We have simply no
information. The training set almost certainly includes code with 
these features, so it is conceivable that TransCoder can do something
with them. But it is a very safe bet that it would do less well on 
more sophisticated code, and a pretty good bet that it would lose some
of its edge over the hand-crafted competition. (One disadvantage of 
computational accuracy as a measure as compared to BLEU is that
producing correct code requires getting everything right; and as
the code gets longer, that becomes increasingly unlikely, even without
introducing more programming language features. BLEU, by contrast,
is more or less scale-invariant.)

The real mystery is why, given the limited nature of the test set, 
the commercial transcompilers
that were used for comparison do so poorly on it.
One conjecture that seems plausible is that the translating the 
``sophisticated'' examples in the test set
requires building in correspondences between library functions in the different
languages, and that the designers of the commercial transcompilers did 
not invest their energies in that aspect of the languages. But I have not
attempted to determine whether that is in fact the correct explanation.

\section{Conclusion}
Despite all that, TransCoder remains a remarkable
accomplishment; it still seems to me very surprising that it works
at all,  even over this limited class of programs. Nor can one fault
the creators of TransCoders for having used this test set; it is not
easy to find a large corpus of parallel programs. However, in 
evaluating the scope and significance of this accomplishment, it is
critical to keep in mind the important limitations of the test set
over which it has been evaluated. TransCoder has received a fair amount
of attention and discussion in the broader community; it is safe to say
that none of this discussion was informed by an understanding
of those limitations. It would have been better if the characteristics
and limitations of the dataset had been laid out when TransCoder
was first announced.

In recent years it has become all too common in AI to generate data sets and
use them in evaluations without sufficiently examining and analyzing 
the actual data that they contain. When a new test set is being introduced,
this is all the more important. 
In one particularly egregious recent instance, the CycIC 
dataset\footnote{https://leaderboard.allenai.org/cycic/submissions/public/}
was created synthetically, and presented to the community as a benchmark,
and tested against human subjects, and posted on the Allen AI leaderboard; 
and twenty systems competed on the leaderboard, with accuracies ranging from 
14\% to 94\%; and their outcomes were presented at prestigious meetings
in Powerpoint slides
--- without, apparently, anyone but me ever noticing that the
data set contained multiple sets of minor variants of the same, ill-designed,
problems repeated literally hundreds of times (Davis, 2020). 
We are all wearyingly familiar with the 
AI research paper full of {\em de rigeur\/}
elaborate 20x20x20 tables of systems with
various forms of ablation vs. datasets
vs. metrics with SOTA in boldface, but not a single concrete 
example.\footnote{Heng Ji (personal communication) estimates that more than 
70\% of the papers at ACL have no examples or qualitative analysis.}
These may be of value to the research teams competing on the specific problem,
but to the wider scientific community, a table of ten examples where the system
succeeds and ten where it fails is much more informative --- and, please,
not just the shopworn standards we have all seen hundreds of times.
Conference and journal reviewers should insist on these.
The gender and racial biases in text and image datasets collected from the web 
and in the programs that have been trained on them
are notorious (Sweeney, 2013), (Buolamwini and Gebru, 2018) (Sun et al., 2019).
If we want to build high-quality AI systems, to understand their scope and
limits, and to explain to the world at large both the significance of the
particular systems and the promise and challenges of AI generally, we
need to carefully examine the datasets we are using to evaluate them.

\section*{Acknowledgements}
Thanks to Thomas Wies and Heng Ji for helpful feedback.

\section*{References}
Joy Buolamwini and Timnit Gebru (2018). 
“Gender shades: Intersectional accuracy disparities in commercial 
gender classification.” {\em Conference on Fairness, Accountability and 
Transparency,} 2018. 77–91. \\
\url{http://proceedings.mlr.press/v81/buolamwini18a.html}

Ernest Davis (2020). ``Question templates in the CycIC training set''.\\
\url{https://cs.nyu.edu/faculty/davise/papers/CYCQns.html}

Marie-Anne Lachaux, Baptiste Roziere, Lowik Chanussot, and 
Guillaume Lample (2020). 
``Unsupervised translation of programming
languages''. arXiv preprint arXiv:2006.03511 \\
\url{https://arxiv.org/abs/2006.03511}

Guillaume Lample, Alexis Conneau, Ludovic Denoyer, and Marc'Aurelio Ranzato. 
(2017).  ``Unsupervised machine translation using monolingual corpora only.'' 
arXiv preprint arXiv:1711.00043.  \\
\url{https://arxiv.org/abs/1711.00043}

Baptiste Roziere, Marie-Anne Lachaux, Lowik Chanussot, and 
Guillaume Lampel (2020). ``Deep learning to translate between programming
languages''. {\em Facebook AI Research blog}. \\
\url{https://ai.facebook.com/blog/deep-learning-to-translate-between-programming-languages/}

Tony Sun et al. (2019). ``Mitigating Gender Bias in Natural Language
Processing: Literature Review''

Latanya Sweeney (2013). “Discrimination in online ad delivery.” 
{\em Queue} {\bf 11}(3): 10. \\
\url{https://arxiv.org/abs/1301.6822.}
\end{document}